\documentclass[letterpaper, 10 pt, conference]{ieeeconf}

\IEEEoverridecommandlockouts                              
\usepackage[colorlinks=true,citecolor=black,linkcolor=black, urlcolor=black]{hyperref}
\usepackage{cite}
\usepackage{amsmath,amssymb,amsfonts}
\usepackage{multirow}
\usepackage{booktabs}
\usepackage{caption}
\usepackage{algorithm}
\usepackage{algpseudocode}
\usepackage{graphicx}
\usepackage{textcomp}
\usepackage{scalerel}
\usepackage[hang,flushmargin]{footmisc} 
\usepackage{siunitx}
\usepackage[update]{epstopdf}
\usepackage[process=all,crop=pdfcrop,cleanup={.dvi,.ps,.pdf,.log,.aux,.bbl}]{pstool}
\usepackage{addfont}
\addfont{OT1}{rsfs10}{\rsfs}

\usepackage{amsthm}

\algblockdefx{MRepeat}{EndRepeat}{\textbf{repeat}}{}
\algnotext{EndRepeat}
\algblockdefx{MFor}{EndFor}{\textbf{for}}{\textbf{do}}
\algnotext{EndFor}
\algnotext{EndIf}

\makeatletter
\let\OldStatex\Statex
\renewcommand{\Statex}[1][3]{%
  \setlength\@tempdima{\algorithmicindent}%
  \OldStatex\hskip\dimexpr#1\@tempdima\relax}
\makeatother

\DeclareMathOperator*{\argmin}{\text{argmin}}

\title{\LARGE \bf Real-Time Capable Decision Making \\ for Autonomous Driving Using Reachable Sets}

\author{Niklas Kochdumper and Stanley Bak
\thanks{The authors are with the Department of Computer Science, Stony Brook University, NY, USA (e-mail: \{\texttt{niklas.kochdumper, stanley.bak\}@stonybrook.edu}). }}

\begin{document}

\captionsetup[figure]{
    width=\linewidth, 
    labelfont=bf,        
    font=small,          
}

\newcommand{\operator}[1]{{\normalfont \texttt{#1}}}
\newcommand{\points}{s}
\newcommand{\koop}{r}
\newcommand{\dist}{p}
\newcommand{\gens}{q}
\newcommand{\numobs}{o}
\newcommand{\R}{\mathbb{R}}
\newcommand{\state}{z}

\newcommand{\cmmnt}[1]{\ignorespaces}

\newtheoremstyle{style}
  {\topsep}
  {\topsep}
  {\itshape}
  {0pt}
  {\bfseries}
  {:}
  { }
  {\thmname{#1}\thmnumber{ #2}\thmnote{ (#3)}}

\theoremstyle{style}
\newtheorem{definition}{Definition}
\newtheorem{proposition}{Proposition}
\newtheorem{theorem}{Theorem}
\newtheorem{lemma}{Lemma}
\newtheorem{remark}{Remark}
\newtheorem{assumption}{Assumption}
\newtheorem{example}{Example}
\newtheorem{corollary}{Corollary}
\newtheorem{problem}{Problem}

\setlength{\textfloatsep}{0.1cm}

\maketitle
\thispagestyle{empty}
\pagestyle{empty}

\begin{abstract}
	Despite large advances in recent years, real-time capable motion planning for autonomous road vehicles remains a huge challenge. 
	In this work, we present a decision module that is based on set-based reachability analysis: First, we identify all possible driving corridors by computing the reachable set for the longitudinal position of the vehicle along the lanelets of the road network, where lane changes are modeled as discrete events. Next, we select the best driving corridor based on a cost function that penalizes lane changes and deviations from a desired velocity profile. 
Finally, we generate a reference trajectory inside the selected driving corridor, which can be used to guide or warm start low-level trajectory planners. 
For the numerical evaluation we combine our decision module with a motion-primitive-based and an optimization-based planner and evaluate the performance on 2000 challenging CommonRoad traffic scenarios as well in the realistic CARLA simulator. The results demonstrate that our decision module is real-time capable and yields significant speed-ups compared to executing a motion planner standalone without a decision module.
\end{abstract}



\section{Introduction} \label{sec:introduction}

A typical architecture for an autonomous driving system consists of a navigation module that plans a route (e.g. a sequence of roads that lead to the destination), a decision module that makes high-level choices like when to do a lane change or overtake another car, a motion planning module that constructs a collision-free and dynamically feasible trajectory, and a controller that counteracts disturbances like wind, model uncertainty, or a slippery road to keep the car on the planned trajectory. This paper presents a novel approach for decision making, which is based on reachable sets.



\begin{figure}[!tb] 
	\centering
	\includegraphics[width = 0.45\textwidth]{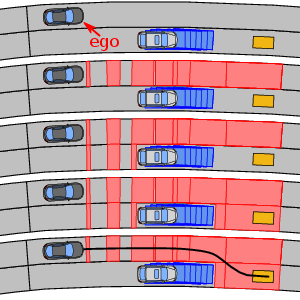}
	\caption{Steps for our decision making approach for an exemplary traffic scenario, where the drivable area is depicted in red, the space occupied by the other traffic participants in blue, the goal set in yellow, and the generated reference trajectory in black.}
	\label{fig:trafficScenario}
\end{figure}

\subsection{State of the Art}

Let us first review the state of the art for decision making and motion planning for autonomous road vehicles. Motion planning approaches can be classified into the four groups \textit{graph search planners}, \textit{sampling-based planners}, \textit{optimization-based planners}, and \textit{interpolating curve planners}. Graph search planners \cite{Anderson2012,Kala2013,Ferguson2008b,Dolgov2010,Ziegler2009,McNaughton2011} represent the search space by a finite grid, where the grid cells represent the nodes and transitions between grid cells the edges of a graph. Motion planning then reduces to the task of finding the optimal path through the graph, which can be efficiently realized using graph search algorithms such as Dijkstra \cite{Anderson2012,Kala2013} and A* \cite{Ferguson2008b,Dolgov2010}. The main disadvantage of this approach is the often large number of grid cells required to cover the search space, especially if spatio-temporal lattices \cite{Ziegler2009,McNaughton2011} are used as a grid.
While for graph search planners the discrete motion primitives are deterministically defined by the grid, sampling-based planners choose motions randomly to explore the search space. For autonomous driving, this is often implemented using rapidly exploring random trees \cite{Kuwata2009,Karaman2011b,Kala2011}. Disadvantages are that sampling based planners in general do not find the optimal solution in finite time, and that the determined trajectories are often jerky and therefore have low driving comfort. Optimization-based planners \cite{Qian2016,Miller2018a,Rosolia2016,Zhang2020,Carvalho2013,Schulman2014} determine trajectories by minimizing a specific cost function with respect to the constraints of dynamic feasibility with the vehicle model and avoiding collisions with other traffic participants and the road boundary. The main challenge is to incorporate the non-convex collision avoidance constraints, which is often realized using mixed-integer programming \cite{Qian2016,Miller2018a}, nonlinear programming with a suitable initial guess \cite{Rosolia2016,Zhang2020}, or via successive convexification \cite{Carvalho2013,Schulman2014}. However, all these methods are either computationally expensive or have the risk of getting stuck in local minima. Finally, interpolating curve planners construct trajectories by interpolating between a sequence of desired waypoints using clothoids \cite{Brezak2013,Coombs2000}, polynomial curves \cite{Piazzi2002,Lee2012b}, Bezier curves \cite{Montes2008,Liang2012}, or splines \cite{Berglund2009,Labakhua2006}. Obstacle avoidance can be realized by modifying the waypoints accordingly \cite{Gonzalez2014,Han2010}.
While interpolating curve planners are able to create smooth trajectories with high comfort, the solutions might not be optimal with respect to a given cost function. A more detailed overview of different motion planning approaches is provided in \cite{Gonzalez2015,Paden2016}.

Even though many of the above motion planners already have the ability to make decisions, it is still a common practice to separate high-level decision making from low-level trajectory planning since this usually simplifies the motion planning problems and therefore improves computational efficiency. One frequently applied method is rule-based decision making, which is often realized using state-machines \cite{Kammel2008,Bacha2008}. Another approach is topological trajectory grouping, which selects topological patterns from a pool of trajectories \cite{Gu2016,Esterle2018}. Also hybrid automata can be used for decision making, where the discrete modes of the automaton represent the high-level decisions \cite{Ahn2020}. Yet another strategy is to apply reinforcement learning for high-level decision making \cite{Mirchevska2018,Mukadam2017}. Finally, some recent approaches \cite{Manzinger2020,Schafer2023,Wuersching2021} use set-based reachability analysis to identify potential driving corridors. 
These methods linearize the system around a given reference path to construct the drivable area by computing the reachable set in longitudinal and lateral direction. However, this has the disadvantage that the linearization can become very inaccurate or conservative if the vehicle significantly deviates from the reference path. Our approach avoids this problem since we only compute the reachable set for the longitudinal position along the lanelet, and model changes in lateral position as discrete events.

\subsection{Notation}

We denote vectors and scalars $a \in \mathbb{R}^n$ by lowercase letters, matrices $M \in \mathbb{R}^{m \times n}$ by uppercase letters, sets $\mathcal{S} \subset \mathbb{R}^{n}$ by calligraphic letters, and lists $\mathbf{L} = (L_1, \dots L_n)$ by bold uppercase letters. Given two matrices $M_1 \in \mathbb{R}^{n \times m}$ and $M_2 \in \mathbb{R}^{n \times w}$, $[M_1 ~ M_2] \in \mathbb{R}^{n \times m+w}$ denotes their concatenation. Empty sets and lists are denoted by $\emptyset$, and we assume that all lists are ordered. Given sets $\mathcal{S}_1,\mathcal{S}_2 \subset \mathbb{R}^n$ and a matrix $M \in \mathbb{R}^{m \times n}$, we require the set operations linear map $M \, \mathcal{S}_1 := \{ M \, s~|~ s \in \mathcal{S}_1 \}$, Minkowski sum $\mathcal{S}_1 \oplus \mathcal{S}_2 := \{ s_1 + s_2 \, | \, s_1 \in \mathcal{S}_1,\,s_2 \in \mathcal{S}_2\}$, Cartesian product $\mathcal{S}_1 \times \mathcal{S}_2 := \{ [s_1^T~s_2^T]^T \, | \, s_1 \in \mathcal{S}_1, \, s_2 \in \mathcal{S}_2 \}$, intersection $\mathcal{S}_1 \cap \mathcal{S}_2 := \{ s \, | \, s \in \mathcal{S}_1 \wedge s \in \mathcal{S}_2 \}$, union $\mathcal{S}_1 \cup \mathcal{S}_2 := \{ s \, | \, s \in \mathcal{S}_1 \vee s \in \mathcal{S}_2 \}$, and set difference $\mathcal{S}_1 \setminus \mathcal{S}_2 := \{ s \, |\, s \in \mathcal{S}_1 \wedge s \not \in \mathcal{S}_2 \}$. We represent one-dimensional sets by intervals $\mathcal{I} = [\underline{i},\overline{i}] := \{x \, | \, \underline{i} \leq x \leq \overline{i}\}$ and two-dimensional sets by polygons, for which all of the above set operations can be performed efficiently.

\section{Problem Formulation}

We represent the dynamics of the vehicle with a kinematic single track model \cite[Chapter~2.2]{Rajamani2012}:
\begin{equation}
	\begin{split}
		\dot x &= v \, \cos( \varphi ) \quad \quad \dot v = a \\
		\dot y &= v \, \sin( \varphi ) \quad \quad \dot \varphi = \frac{v}{\ell_{wb}} \, \tan( s ),
	\end{split}
	\label{eq:dynamics}
\end{equation}
where the vehicle state consists of the position of the rear axis represented by $x$, $y$, the velocity $v$, and the orientation $\varphi$. The parameter $\ell_{\text{wb}}$ is the length of the vehicle's wheelbase, and the control inputs are the acceleration $a$ and the steering angle $s$, which are bounded by $a \in [-a_{\text{max}}, a_{\text{max}}]$ and $s \in [-s_{\text{max}}, s_{\text{max}}]$. Moreover, an additional constraint is given by the friction circle \cite[Chapter~13]{Rajamani2012}
\begin{equation}
	\sqrt{a^2 + (v \dot \varphi)^2} \leq a_{\text{max}},
	\label{eq:KammsCircle}
\end{equation}
which models the maximum force that can be transmitted by the tires.

The road network is represented by a list of lanelets $\mathbf{L}= (\text{\rsfs L}_1, \dots, \text{\rsfs L}_q)$, where each lanelet is a tuple $\text{\rsfs L} = (\text{id}, \text{left},\text{right},\mathbf{S},\ell_{\text{lane}},\mathcal{L})$ consisting of the lanelet identifier $\text{id} \in \mathbb{N}_{\geq 0}$, the identifiers of the neighboring lanelets to the left and right, $\text{left},\text{right} \in \mathbb{N}_{\geq 0}$, a list $\mathbf{S}$ storing the identifiers of the successor lanelets, the length $\ell_{\text{lane}} \in \mathbb{R}_{\geq 0}$ of the lanelet, and a polygon $\mathcal{L} \subset \mathbb{R}^2$ representing the shape of the lanelet. In addition to the road network, one also has to consider the other traffic participants such as cars or pedestrians for motion planning. We denote the space occupied by traffic participant $j$ at time $t$ by $\mathcal{O}_j(t) \subset \mathbb{R}^2$. For simplicity, we assume that $\mathcal{O}_j(t)$ is known for all surrounding traffic participants. In practice, the current positions of the surrounding traffic participants are obtained via perception using computer vision \cite{Gupta2018}, LiDAR \cite{Sualeh2019}, radar \cite{Xu2020}, or combinations \cite{Jahromi2019, Bertozzi2008}, and the future positions of the traffic participants can be determined using probabilistic \cite{Schulz2018} or set-based prediction \cite{Koschi2020}. 


A motion planning problem is defined by an initial state $x_0$, $y_0$, $v_0$, $\varphi_0$ for the vehicle and a goal region {\rsfs G} $=(\mathcal{G}, \tau_{goal})$ consisting of a goal set for the vehicle state $[x~y~v~\varphi] \in \mathcal{G} \subseteq \mathbb{R}^4$ and a time interval $\tau_{\text{goal}} = [t_{\text{start}}, t_{\text{end}}]$ at which this goal set should be reached. An exemplary motion planning problem is visualized at the top of Fig.~\ref{fig:trafficScenario}. The task for motion planning is to determine control inputs $a(t)$ and $s(t)$ that drive the car from the initial state to the goal region under consideration of the vehicle dynamics in \eqref{eq:dynamics} and such that the vehicle stays on the road and does not collide with other traffic participants at all times. In this paper we propose to solve motion planning problems with a novel decision making module that determines a suitable driving corridor that leads to the goal set and generates a desired reference trajectory inside this driving corridor. Our decision module can then be combined with a low-level trajectory planner that tracks the reference trajectory and generates the control input trajectories $a(t)$ and $s(t)$.


\begin{algorithm}[!tb]
	\caption{Compute drivable area for a single lanelet} \label{alg:driveArea}
	{\raggedright \textbf{Input:} Lanelet $\text{\rsfs L} = (\text{id}, \text{left}, \text{right}, \mathbf{S}, \ell_{\text{lane}}, \mathcal{L})$, list of initial sets $\mathbf{X} = (\mathcal{X}(t_{i_\text{init}}), \dots, \mathcal{X}(t_{i_{\text{fin}}}))$, space occupied by the other traffic participants $\mathcal{O}_1(t), \dots, \mathcal{O}_{\numobs}(t)$ \vspace{3pt}
		
	\textbf{Output:} Drivable area $\mathbf{D} = (\mathcal{D}(t_{i_\text{init}}), \dots, \mathcal{D}(t_{i_{\text{end}}}))$, lists $\mathbf{T}_{\text{left}}, \mathbf{T}_{\text{right}}, \mathbf{T}_{\text{suc}}$ with possible transitions to the left, right, and successor lanelets
	} \vspace{3pt}
	\begin{algorithmic}[1]
		\State $\mathcal{D}(t_{i_\text{init}}) \gets $ set $\mathcal{X}(t_{i_\text{init}})$ with the smallest time from $\mathbf{X}$ 
		\State $\mathbf{T}_{\text{left}}, \mathbf{T}_{\text{right}}, \mathbf{T}_{\text{suc}} \gets \emptyset$
		\State $\mathbf{F}(t) \gets $ compute free space on current, left, right, and 
		\Statex[1] ~~~ successor lanelets from  $\mathcal{O}_1(t), \dots, \mathcal{O}_{\numobs}(t)$ (see \eqref{eq:freeSpace}) \vspace{-6pt}
		\Statex[0] // loop over all time steps
		\For{$i \gets i_\text{init}$ to $\lceil t_{\text{end}} / \Delta t \rceil$} \vspace{3pt}
			\Statex[1] // compute reachable set at next time step (see \eqref{eq:reachableSet})
			\State $\mathcal{D}(t_{i+1}) \gets A \, \mathcal{D}(t_{i}) \oplus B \, [-a_{\text{max}},a_{\text{max}}]$ \vspace{3pt}
			\Statex[1] // unite with initial set for this time step
			\If{$\exists \mathcal{X}(t_{i+1}) \in \mathbf{X}$}
				\State $\mathcal{D}(t_{i+1}) \gets \mathcal{D}(t_{i+1}) \cup \mathcal{X}(t_{i+1})$
			\EndIf
			\Statex[1] // intersect with free space on the lanelet
			\For{$\mathcal{F} \in \mathbf{F}_{\text{id}}(t_{i+1})$}
				\If{$\mathcal{F} \cap \mathcal{D}(t_{i+1}) \neq \emptyset$}
					\State $\mathcal{D}(t_{i+1}) \gets \mathcal{D}(t_{i+1}) \cap \mathcal{F}$
				\EndIf
			\EndFor
			\Statex[1] // terminate if drivable area is empty
			\If{$\mathcal{D}(t_{i+1}) = \emptyset$}
				\State \textbf{break}
			\EndIf
			\Statex[1] // check if a lane change to the right is possible
			\For{$\mathcal{F} \in \mathbf{F}_{\text{right}}(t_{i+1})$}
				\If{$\mathcal{D}(t_{i+1}) \cap \mathcal{F} \neq \emptyset$}
					\State $\mathbf{T}_{\text{right}} \gets$ add set $\mathcal{D}(t_{i+1}) \cap \mathcal{F}$ to $\mathbf{T}_{\text{right}}$
				\EndIf
			\EndFor
			\Statex[1] // check if a lane change to the left is possible
			\For{$\mathcal{F} \in \mathbf{F}_{\text{left}}(t_{i+1})$}
				\If{$\mathcal{D}(t_{i+1}) \cap \mathcal{F} \neq \emptyset$}
					\State $\mathbf{T}_{\text{left}} \gets$ add set $\mathcal{D}(t_{i+1}) \cap \mathcal{F}$ to $\mathbf{T}_{\text{left}}$
				\EndIf
			\EndFor
			\Statex[1] // check if a lane change to a successor is possible
			\State $\mathcal{D}_{\text{shift}} \gets \mathcal{D}(t_{i+1}) \oplus [-\ell_{\text{lane}}~0]^T$
			\For{$\text{suc} \in \mathbf{S}$}
				\For{$\mathcal{F} \in \mathbf{F}_{\text{suc}}(t_{i+1})$}
					\If{$\mathcal{D}_{\text{shift}} \cap \mathcal{F} \neq \emptyset$}
						\State $\mathbf{T}_{\text{suc}} \gets$ add set $\mathcal{D}_{\text{shift}} \cap \mathcal{F}$ to $\mathbf{T}_{\text{suc}}$
					\EndIf
				\EndFor
			\EndFor
		\EndFor
		\Statex[0] // combine sets for successive steps into one transition 
		\State $\mathbf{T}_{\text{left}}, \mathbf{T}_{\text{right}}, \mathbf{T}_{\text{suc}} \gets $ group successive sets together
	\end{algorithmic}
\end{algorithm}

\section{Algorithm}
\label{sec:algorithm}

We now present our novel approach for decision making. 

\subsection{Simplified Vehicle Dynamics}


For the sake of computational efficiency, it is common practice to use a simplified version of the vehicle dynamics in \eqref{eq:dynamics} for decision making. To obtain such a simplified model, we introduce a curvilinear coordinate frame that follows the lanelet centerline, and in which $\xi$ denotes the longitudinal position of the vehicles center along the corresponding lanelet. Moreover, we represent the vehicle as a discrete-time system with time step size $\Delta t$, where $t_i = i \, \Delta t$ are the corresponding time points and we assume without loss of generality that the initial time is $t_0 = 0$. Our simplified model for decision making is then given by the following double integrator for the longitudinal position $\xi$
\begin{equation}
	\begin{bmatrix} \xi(t_{i+1}) \\ v(t_{i+1}) \end{bmatrix} = \underbrace{\begin{bmatrix} 1 & \Delta t \\ 0 & 1\end{bmatrix}}_{A} \begin{bmatrix} \xi(t_i) \\ v(t_i) \end{bmatrix} + \underbrace{\begin{bmatrix} 0.5 \, \Delta t^2 \\ \Delta t \end{bmatrix}}_{B} a_i,
	\label{eq:doubleIntegratorDiscrete}
\end{equation}
where $v$ is the velocity and $a_i$ is the constant acceleration of the vehicle during time step $i$. In the remainder of the paper we will use notation $\state(t) = [\xi(t) ~ v(t)]^T$ for the state of the simplified vehicle dynamics in \eqref{eq:doubleIntegratorDiscrete}.

\subsection{Drivable Area}

To determine all possible driving corridors, we use reachability analysis. The reachable set $\mathcal{R}(t)$ for system \eqref{eq:doubleIntegratorDiscrete} is defined as the set of longitudinal positions and velocities reachable under consideration of the bounded acceleration $[-a_{\text{max}},a_{\text{max}}]$. This set can be computed by the following propagation rule:
\begin{equation}
	\mathcal{R}(t_{i+1}) = A \, \mathcal{R}(t_i) \oplus B \, [-a_{\text{max}},a_{\text{max}}],
	\label{eq:reachableSet}
\end{equation}
where we represent sets by polygons. To avoid collisions with other traffic participants, we have to consider their occupied space $\mathcal{O}_j(t)$. Given a lanelet $\text{\rsfs L} = (\text{id}, \text{left},\text{right},\mathbf{S},\ell_{\text{lane}},\mathcal{L})$, we therefore first compute the set $\mathcal{O}_{\text{long},\text{id},j}(t) \subset \mathbb{R}$ of longitudinal lanelet positions occupied by obstacle $j$ at time $t_i$ from the occupied space $\mathcal{O}_j(t_i)$ in the global coordinate frame. Using this set, we can compute the free space on the lanelet as follows:
\begin{equation}
\begin{split}
	\mathcal{F}_{\text{id}}(t_i) = & \bigg( [0, \ell_{\text{lane}}] \setminus \bigg( \bigcup_{j=1}^{\numobs} \mathcal{O}_{\text{long},\text{id},j}(t_i) ~ \oplus \\
	& \Big[-\frac{\ell_{\text{car}}}{2} - d_{\text{min}}, \frac{\ell_{\text{car}}}{2} + d_{\text{min}}\Big] \bigg)\bigg) \hspace{-3pt} \times [0, v_{\text{max},\text{id}}],
	\label{eq:freeSpace}
	\end{split}
\end{equation}
where we subtract the occupied space for all $\numobs$ traffic participants and bloat the obstacles by the length of the ego vehicle $\ell_{\text{car}}$ as well as by a user-defined minimum distance $d_{\text{min}}$ we want to keep to the other traffic participants. In addition, we take the Cartesian product with the set of legal velocities $[0, v_{\text{max},\text{id}}]$, where $v_{\text{max},\text{id}}$ is the speed limit for the current lanelet. Since the free space $\mathcal{F}_{\text{id}}(t_i)$ in general consists of multiple disjoint regions, we introduce the list $\mathbf{F}_{\text{id}}(t_i)$ that stores all these disjoint regions. The drivable area is finally given by the intersection of the reachable set with the free space on the lanelet:
\begin{equation}
	\mathcal{D}(t_i) = \mathcal{R}(t_i) \cap \mathcal{F}_{\text{id}}(t_i).
	\label{eq:driveArea}
\end{equation}
While equations \eqref{eq:reachableSet}, \eqref{eq:freeSpace}, \eqref{eq:driveArea} enable us to compute the drivable area for a single lanelet, we additionally have to consider transitions between the lanelets to obtain the drivable area for the whole road network. This procedure is summarized in Alg.~\ref{alg:driveArea}, which computes the drivable area for a single lanelet under consideration of transitions. The algorithm takes as input a list of drivable areas that correspond to transitions to the given lanelet and computes the drivable area for the lanelet as well as all possible transitions to left, right, or successor lanelets.

\subsection{Driving Corridor Selection}

Using the approach for computing the drivable area for a single lanelet together with possible transitions to other lanelets in Alg.~\ref{alg:driveArea}, we can formulate the identification of possible driving corridors as a standard tree search problem, where each node consists of a lanelet and a corresponding drivable area. This procedure is summarized in Alg.~\ref{alg:main} and visualized for an exemplary traffic scenario in Fig.~\ref{fig:trafficScenario}. Once we identified all driving corridors that reach the goal set, we finally have to select the best driving corridor in Line~\ref{line:selection} of Alg.~\ref{alg:main}. For this, we use the cost function
\begin{equation}
	J = w_{\text{change}} \, n_{\text{change}} + w_{\text{profile}} \, d_{\text{profile}},
	\label{eq:cost}
\end{equation}
where $n_{\text{change}}$ is the number of lane changes for the driving corridor, $d_{\text{profile}}$ is the average deviation from a desired position-velocity-profile $\state_{\text{des}}(t)$, and $w_{\text{change}}, w_{\text{profile}} \in \mathbb{R}_{\geq 0}$ are user-defined weighting factors. For the desired position-velocity-profile, we choose to accelerate to the current speed limit $v_{\text{max},\text{id}} $ with a user-defined desired acceleration $a_{\text{des}}$:
\begin{equation*}
	\state_{\text{des}}(t_{i+1}) = A \, \state_{\text{des}}(t_{i}) + B \, a_i
\end{equation*}
with $\state(t_0) = [\xi_0~v_0]^T$ and
\begin{equation*}
	a_i = \max \Big(-a_{\text{des}}, \min \Big(a_{\text{des}}, \frac{v_{\text{max},\text{id}} - v(t_i)}{\Delta t} \Big) \Big).
\end{equation*}
The deviation $d_{\text{profile}}$ is given as the minimum deviation inside the drivable area $\mathcal{D}(t_i)$ averaged over all time steps:  
\begin{equation*}
	d_{\text{profile}} = \frac{1}{\lceil t_{\text{end}}/\Delta t \rceil} \sum_{i=0}^{\lceil t_{\text{end}} / \Delta t\rceil} \min_{\state \in \mathcal{D}(t_i)} \| \state - \state_{\text{des}}(t_i) \|_2.
\end{equation*}
If the driving corridor contains multiple drivable areas on different lanelets for the same time step, we take the minimum deviation from all these areas. Moreover, we shift the desired position-velocity-profile by the lanelet length $\ell_{\text{lane}}$ when moving on to a successor lanelet.  

\begin{algorithm}[!tb]
	\caption{Determine best driving corridor} \label{alg:main}
	{\raggedright \textbf{Input:} Initial state $x_0$, $y_0$, $v_0$, $\varphi_0$ of the vehicle, road network given as a list of lanelets $\mathbf{L}= (\text{\rsfs L}_1, \dots, \text{\rsfs L}_q)$, goal region {\rsfs G} $=(\mathcal{G}, \tau_{\text{goal}})$, space occupied by the other traffic participants $\mathcal{O}_1(t), \dots, \mathcal{O}_{\numobs}(t)$ \vspace{3pt}
		
	\textbf{Output:} Best driving corridor given by a sequence of lanelets $\mathbf{L}_{\text{fin}} = (\text{\rsfs L}_1, \dots,\text{\rsfs L}_p)$ and the corresponding drivable areas $\mathbf{D}_{\text{fin}} = (\mathbf{D}_1, \dots, \mathbf{D}_p)$, with $\mathbf{D}_k = (\mathcal{D}(t_{k_\text{init}}), \dots \mathcal{D}(t_{k_{\text{end}}}))$
	} \vspace{3pt}
	\begin{algorithmic}[1]
	\State $\text{\rsfs L}_0 \gets$ find lanelet for the initial state $x_0$, $y_0$, $\varphi_0$
	\State $\xi_0 \gets$ transform $x_0$, $y_0$ into curvilinear coordinate system
	\State $\text{\rsfs L}_{\text{goal}} \gets$ find lanelet for the goal region $\mathcal{G}$
	\State $\mathcal{G}_{\text{long}} \gets$ transform $\mathcal{G}$ into curvilinear coordinate system
	\State $\mathbf{Q} \gets $ initialize queue with set $[\xi_0~ v_0]^T$ and lanelet $\text{\rsfs L}_0$
	\Repeat
		\State $\mathbf{X}, \text{\rsfs L} \gets$ list of initial sets and the corresponding 
		\Statex[3] ~~lanelet for first element from the queue $\mathbf{Q}$
		\State $\mathbf{D}, \mathbf{T}_{\text{left}}, \mathbf{T}_{\text{right}}, \mathbf{T}_{\text{suc}} \gets$ compute drivable area for 
		\Statex[4] lanelet $\text{\rsfs L}$ starting from $\mathbf{X}$ using Alg.~\ref{alg:driveArea}
		\State $\mathbf{T}_{\text{left}}, \mathbf{T}_{\text{right}}, \mathbf{T}_{\text{suc}} \gets $ remove entries that are already 
		\Statex[5] covered by an existing drivable area 
		\State $\mathbf{Q} \gets$ add entries in $\mathbf{T}_{\text{left}}, \mathbf{T}_{\text{right}}, \mathbf{T}_{\text{suc}}$ and the 
		\Statex[3] corresponding lanelets to the queue $\mathbf{Q}$
		\For{$\mathcal{D}(t_i) \in \mathbf{D}$}
			\If{$\text{\rsfs L} = \text{\rsfs L}_{\text{goal}} \wedge t_i \in \tau_{\text{goal}} \wedge \mathcal{D}(t_i) \subseteq \mathcal{G}_{\text{long}}$}
				\State $\mathbf{F}_{\text{fin}} \gets $ add current driving corridor to $\mathbf{F}_{\text{fin}}$
			\EndIf
		\EndFor
	\Until{$\mathbf{Q} = \emptyset$}
	\State $\mathbf{L}_{\text{fin}}, \mathbf{D}_{\text{fin}} \gets$ select driving corridor with the lowest  
	\Statex[3] $~~$ cost according to \eqref{eq:cost} from $\mathbf{F}_{\text{fin}}$ \label{line:selection}
	\end{algorithmic}
\end{algorithm}

\begin{figure}[!tb] 
	\centering
	\includegraphics[width = 0.48\textwidth]{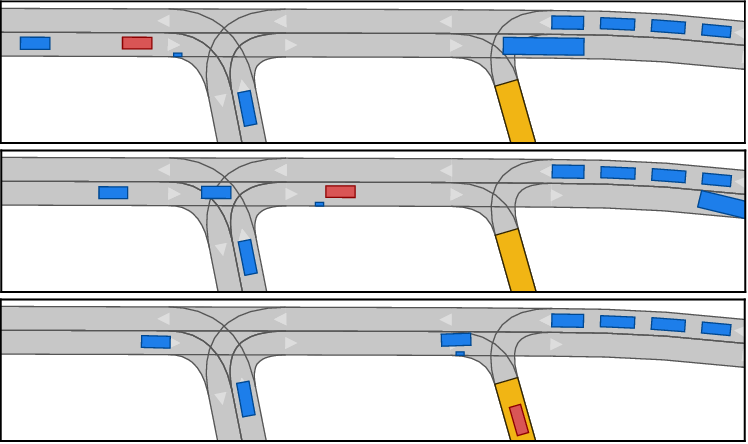}
	\caption{Trajectory planned by our decision module in combination with the optimization-based planner for the CommonRoad scenario DEU\_Flensburg-73\_1\_T-1 visualized at times 0\si{\second}, 3\si{\second}, and 6\si{\second}, where the ego vehicle is shown in red, the other traffic participants in blue, and the goal set in yellow.}
	\label{fig:commonRoad}
\end{figure}

\subsection{Driving Corridor Refinement}

While the driving corridor computed using Alg.~\ref{alg:main} is guaranteed to contain at least one state sequence that leads to the goal set, it usually also contains states that do not reach the goal. To remove those states, we refine the computed driving corridor by propagating the reachable sets backward in time starting from the intersection of the final set $\mathcal{D}(t_{\text{end}})$ with the goal set $\mathcal{G}_{\text{long}}$ from Alg.~\ref{alg:main}. The refined drivable area is then given by the intersection of the backpropagated sets with the original drivable area, which corresponds to the following propagation rule: 
  \begin{equation*}
  	\mathcal{D}(t_{i-1}) = \mathcal{D}(t_{i-1}) \cap A^{-1} \big( \mathcal{D}(t_{i}) \oplus B \, [-a_{\text{max}}, a_{\text{max}}] \big),
  \end{equation*}
  where $\mathcal{D}(t_{\text{end}}) = \mathcal{D}(t_{\text{end}}) \cap \mathcal{G}_{\text{long}}$. Again, we shift the drivable area by the lanelet length $\ell_{\text{lane}}$ when moving on to a predecessor lanelet.

\begin{table*}
\renewcommand{\arraystretch}{1.2}
\begin{center}
\caption{Performance of different motion planners for 2000 CommonRoad traffic scenarios, where the evaluation metrics are the average computation time in milliseconds for planning a trajectory with a duration of one second, the percentage of successfully solved scenarios, and the percentage of scenarios where the vehicle collides with other cars or leaves the road.}
\vspace{-3pt}
\label{tab:commonRoad}
\begin{tabular}{ l c c c c c c c c c c c c c c}
 \toprule
 \multirow{3}{1.2cm}{ \\ \textbf{Scenarios} \\} & \multicolumn{2}{c}{\textbf{Decision Module}} & & \multicolumn{3}{c}{\textbf{Motion Primitives}} & & \multicolumn{3}{c}{\textbf{Decision Module +}} & & \multicolumn{3}{c}{\textbf{Decision Module +}}\\ 
 & \multicolumn{2}{c}{\textbf{Standalone}} & & \multicolumn{3}{c}{\textbf{Standalone}} & & \multicolumn{3}{c}{\textbf{Motion Primitives}} & & \multicolumn{3}{c}{\textbf{Optimization}} \\ 
 \cmidrule{2-3} \cmidrule{5-7} \cmidrule{9-11} \cmidrule{13-15} & time & solved & ~~~ & time & solved & collisions & ~~~ & time & solved & collisions & ~~~ & time & solved & collisions\\ \midrule 
overall & 106 & 100\si{\percent}  & &  4570 & 11\si{\percent} & 0\si{\percent}  & &  302 & 60\si{\percent} & 0\si{\percent}  & &  424 & 91\si{\percent} & 7\si{\percent} \\
$t_{\text{end}} \leq 4\si{\second}$ & 79 & 100\si{\percent}  & &  8324 & 9\si{\percent} & 0\si{\percent} & &  192 & 76\si{\percent} & 0\si{\percent}  & &  343 & 93\si{\percent} & 7\si{\percent} \\
$t_{\text{end}} > 4\si{\second}$ & 139 & 100\si{\percent}  & &  1353 & 14\si{\percent} & 0\si{\percent} & &  581 & 38\si{\percent} & 0\si{\percent}  & &  451 & 92\si{\percent} & 7\si{\percent} \\		
urban & 121 & 100\si{\percent}  & &  2288 & 5\si{\percent} & 0\si{\percent}  & &  506 & 30\si{\percent} & 0\si{\percent}  & &  451 & 92\si{\percent} & 7\si{\percent} \\
highway & 226 & 100\si{\percent}  & &  894  & 48\si{\percent} & 0\si{\percent}  & &  384 & 74\si{\percent} & 0\si{\percent} & &  621 & 89\si{\percent} & 3\si{\percent} \\
intersections~~~~~~ & 98 & 100\si{\percent}  & & 6875 & 7\si{\percent} & 0\si{\percent}  & &  300 & 58\si{\percent} & 0\si{\percent}  & &  413 & 91\si{\percent} & 7\si{\percent} \\				 
 \bottomrule 
\end{tabular}
\end{center}
\vspace{-17pt}
\end{table*}

\subsection{Reference Trajectory Generation}

As the last step of our approach, we generate a suitable reference trajectory inside the selected driving corridor. Optimally, we would like to drive with the desired position-velocity-profile $\state_{\text{des}}(t)$. However, this position-velocity-profile might not be located inside the driving corridor. We therefore generate the reference trajectory by choosing for each time step the point $\state(t_{i+1})$ inside the drivable area $\mathcal{D}(t_{i+1})$ that is closest to $\state_{\text{des}}(t_{i+1})$ and reachable from the previous state $\state(t_i)$ given the dynamics in \eqref{eq:doubleIntegratorDiscrete}: 
\begin{equation*}
	\state(t_{i+1}) = \argmin_{\state \in \mathcal{I}} \| \state - \state_{\text{des}}(t_{i+1}) \|_2
\end{equation*}

\vspace{-10pt}

\noindent where
\begin{equation*}
	\mathcal{I} = \mathcal{D}(t_{i+1}) \cap \big( A \, \state(t_i) \oplus B \, [-a_{\text{max}},a_{\text{max}}] \big).
\end{equation*}
After generating the reference trajectory $\state(t)$ in the curvilinear coordinate frame, we have to transform it to the global coordinate frame. Here, we obtain $x(t)$, $y(t)$ and $\varphi(t)$ from the lanelet centerline, and $v(t)$ is directly given by $\state(t)$. At lane changes to a left or right lanelet we interpolate between the lanelet centerlines $x_{1}(t),y_{1}(t)$ and $x_{2}(t),y_{2}(t)$ of the lanelets before and after the lane change as follows:
\begin{equation*}
	\begin{bmatrix} x(t) \\ y(t) \end{bmatrix} = (1-\mu(t)) \begin{bmatrix} x_{1}(t) \\ y_{1}(t) \end{bmatrix} + \mu(t) \begin{bmatrix} x_{2}(t) \\ y_{2}(t) \end{bmatrix}, 
\end{equation*}
with
\begin{equation*}
	\mu(t) = \frac{1}{1 + e^{-10 (\delta(t)- 0.5)}}, ~~ \delta(t) = \frac{(t - t_{\text{init}})}{t_{\text{fin}} - t_{\text{init}}},
\end{equation*}
where $t_{\text{init}}$, $t_{\text{fin}}$ are the start and end time for the lane change. 

\section{Improvements}
\label{sec:improvements}

We now present several improvements for our algorithm.

\subsection{Traffic Rules}

So far, the only traffic rule we considered is the speed limit. However, our approach makes it easy to already incorporate many additional traffic rules on a high-level during driving corridor generation. For example, traffic rules such as traffic lights, no passing rules, or the right-of-way can simply be considered by removing the space that is blocked by the traffic rule from the free space $\mathcal{F}_{\text{id}}(t)$ on the lanelet. Other rules such as keeping a safe distance to the leading vehicle can be considered by adding the corresponding rule violations with a high penalty to the cost function in \eqref{eq:cost}. Incorporating traffic rule violations into the cost function ensures that our decision module can find a solution even if the initial state violates the rule. This can for example happen if another traffic participant performs an illegal cut-in in front of the ego vehicle, which makes it impossible to keep a safe distance at all times.

\subsection{Partially Occupied Lanelets}

Often, other traffic participants only occupy a small part of the lateral space on the lanelet, for example if a bicycle drives on one side of the lane. Classifying the whole lanelet as occupied in those cases would be very conservative and prevent the decision module from finding a feasible solution in many cases. A crucial improvement for the basic algorithm in Sec.~\ref{sec:algorithm} is therefore to check how much of the lateral space is occupied by the other traffic participants, and only remove the parts where the remaining free lateral space is too small to drive on from the free space in \eqref{eq:freeSpace}. We additionally correct the final reference trajectory accordingly to avoid intersections with other traffic participants that only partially occupy the lateral space of the lanelet. 

\subsection{Cornering Speed}
Our algorithm in Sec.~\ref{sec:algorithm} assumes that the vehicle can drive around corners with arbitrary speed, which obviously is a wrong assumption since the vehicle speed in corners is limited by the friction circle in \eqref{eq:KammsCircle}. Therefore, we now derive a formula for the maximum corner speed given the lanelet curvature $\Delta \varphi/ \Delta \xi$, which we use as an artificial speed limit for all lanelets. We assume that we drive the corner with constant velocity ($a = 0$), in which case \eqref{eq:KammsCircle} simplifies to 
\begin{equation}
	(v \dot \varphi)^2 \overset{\eqref{eq:dynamics}}{=} \Big( \frac{v^2}{\ell_{\text{wb}}} \tan(s) \Big)^2 \leq a_{\text{max}}^2.
	\label{eq:corneringSpeed1}
\end{equation}
Additionally assuming a constant steering angle ($\dot s = 0$), we can estimate the change in orientation as 
\begin{equation}
	\Delta \varphi = \int_{t = 0}^{t = \Delta t} \dot \varphi \, dt \overset{\eqref{eq:dynamics}}{=} \frac{v}{\ell_{\text{wb}}} \tan(s) \Delta t = \frac{\Delta \xi}{\ell_{\text{wb}}} \tan(s), 
	\label{eq:corneringSpeed2}
\end{equation}
where we used the estimation $\Delta t \approx \Delta \xi / v$. Combining \eqref{eq:corneringSpeed1} and \eqref{eq:corneringSpeed2} finally yields
\begin{equation}
	v \leq \sqrt{a_{\text{max}} \, \Delta \xi / \Delta \varphi}
\end{equation}
for the maximum corner velocity. In the implementation we compute $\Delta \varphi / \Delta \xi$ for each segment of the lanelet and take the maximum over all segments.

\subsection{Minimum Lane Change Time}

Alg.~\ref{alg:driveArea} assumes that a single time step is sufficient to perform a lane change, which is unrealistic. We therefore now derive a formula that specifies how many time steps are required to perform a lane change. The orientation of the vehicle during the lane change is modeled as a function
\begin{equation}
	\varphi(t) = \begin{cases} 2 \, \frac{\varphi_{\text{peak}}}{t_{\text{fin}}} \, t, & 0 \leq t \leq t_{\text{fin}}/2 \\ 2 \, \varphi_{\text{peak}} - 2 \, \frac{\varphi_{\text{peak}}}{t_{\text{fin}}} \, t, & t_{\text{fin}}/2 < t \leq t_{\text{fin}} \end{cases},
	\label{eq:laneChange1}
\end{equation}
where $t_{\text{fin}}$ is the time required for the lane change. The maximum peak orientation $\varphi_{\text{peak}}$ we can choose is bounded by the friction circle \eqref{eq:KammsCircle}, for which we obtain under the assumption of constant velocity ($a=0$):
\begin{equation}
	(v \dot \varphi(t))^2 \overset{\eqref{eq:laneChange1}}{=} \Big( 2 \, v \frac{\varphi_{\text{peak}}}{t_{\text{fin}}}\Big)^2 \leq a_{\text{max}}^2.
	\label{eq:laneChange2}
\end{equation}
Moreover, with the small angle approximation $\sin(\varphi) \approx \varphi$, we obtain for the change in lateral position of the vehicle
\begin{equation}
	\Delta \eta \overset{\eqref{eq:dynamics}}{=} \int_{t = 0}^{t = t_{\text{fin}}} v \, \varphi(t) \, dt \overset{\eqref{eq:laneChange1}}{=} 0.5 \, v \, \varphi_{\text{peak}} \, t_{\text{fin}}.
	\label{eq:laneChange3}
\end{equation}
Combining \eqref{eq:laneChange2} and \eqref{eq:laneChange3} finally yields 
\begin{equation}
	t_{\text{fin}} \geq \sqrt{4 \, \Delta \eta / a_{\text{max}}}
\end{equation}
for the minimum time required for a lane change, where $\Delta \eta$ is the lateral distance between the lanelet centerlines.

\begin{table}
\renewcommand{\arraystretch}{1.2}
\begin{center}
\caption{Results for the experiments in CARLA, where we specify the average computation time for our decision module to plan a trajectory with a duration of one second.}
\vspace{-3pt}
\label{tab:CARLA}
\begin{tabular}{ c c c c}
 \toprule
 \textbf{Route} & ~~~\textbf{Distance}~~~ & \textbf{Planning Problems} & \textbf{Computation Time} \\	\midrule
 1 & 789\,\si{\meter} & 622 & 74\,\si{\milli\second}\\
 2 & 685\,\si{\meter} & 489 & 73\,\si{\milli\second}\\
 3 & 409\,\si{\meter} & 346 & 74\,\si{\milli\second}\\			 
 \bottomrule 
\end{tabular}
\end{center}
\vspace{-5pt}
\end{table}


\section{Numerical Evaluation}

We implemented our approach in Python, and all computations are carried out on a 3.5GHz Intel Core i9-11900KF processor. Our implementation is publicly available on GitHub\footnote{\url{https://github.com/KochdumperNiklas/MotionPlanner}}, and we published a repeatability package that reproduces the presented results on CodeOcean\footnote{\url{https://codeocean.com/capsule/8454823/tree/v1}}. For the parameter values of our decision module we use $a_{\text{des}} = 1\si{\metre \per \second}$ for the desired acceleration, $d_{\text{min}} = 1\si{\metre}$ for the minimum safe distance, and $w_{\text{change}} = 10$, $w_{\text{profile}} = 1$ for the weights of the cost function \eqref{eq:cost}. Moreover, the time step size is $\Delta t = 0.1\si{\second}$ for CommonRoad and $\Delta t = 0.2\si{\second}$ for CARLA.

\subsection{CommonRoad Scenarios}
\label{subsec:commonRoad}

CommonRoad \cite{Althoff2017a} is a database that contains a large number of challenging motion planning problems for autonomous vehicles, and is therefore well suited to evaluate the performance of our decision module. For the experiments, we combine our decision module with two different types of motion planners, namely a motion-primitive-based planner and an optimization-based planner. To obtain the motion-primitive-based planner we used the AROC toolbox \cite{Kochdumper2021b} to create a maneuver automaton with 12793 motion primitives by applying the generator space control approach \cite{Schuermann2017c}. 
For the optimization-based planner we solve an optimal control problem with the objective to track the reference trajectory generated by our decision module and the constraint of dynamical feasibility with the vehicle model \eqref{eq:dynamics}.
The results for the evaluation on 2000 CommonRoad traffic scenarios are listed in Tab.~\ref{tab:commonRoad}, where we aborted the planning if the computation took longer than one minute. The outcome demonstrates that our decision module standalone on average runs about 10 times faster than real-time and can generate feasible driving corridors for all scenarios. Moreover, while the computation times for running the motion-primitive-based planner standalone are very high and the success rate is consequently quite low due to timeouts, in combination with our decision module the planner is real-time capable and can solve a large number of scenarios, which nicely underscores the benefits of using a decision module. Finally, even though we do not consider any collision avoidance constraints, the optimization based planner still produces collision-free trajectories most of the time, which can be attributed to the good quality of the reference trajectories generated by our decision module. 
The planned trajectory for an exemplary traffic scenario is visualized in Fig.~\ref{fig:commonRoad}, where the ego vehicle overtakes a bicycle that drives at the side of the road. 

\subsection{CARLA Simulator}

\begin{figure}[!tb] 
	\centering
	\includegraphics[width = 0.48\textwidth]{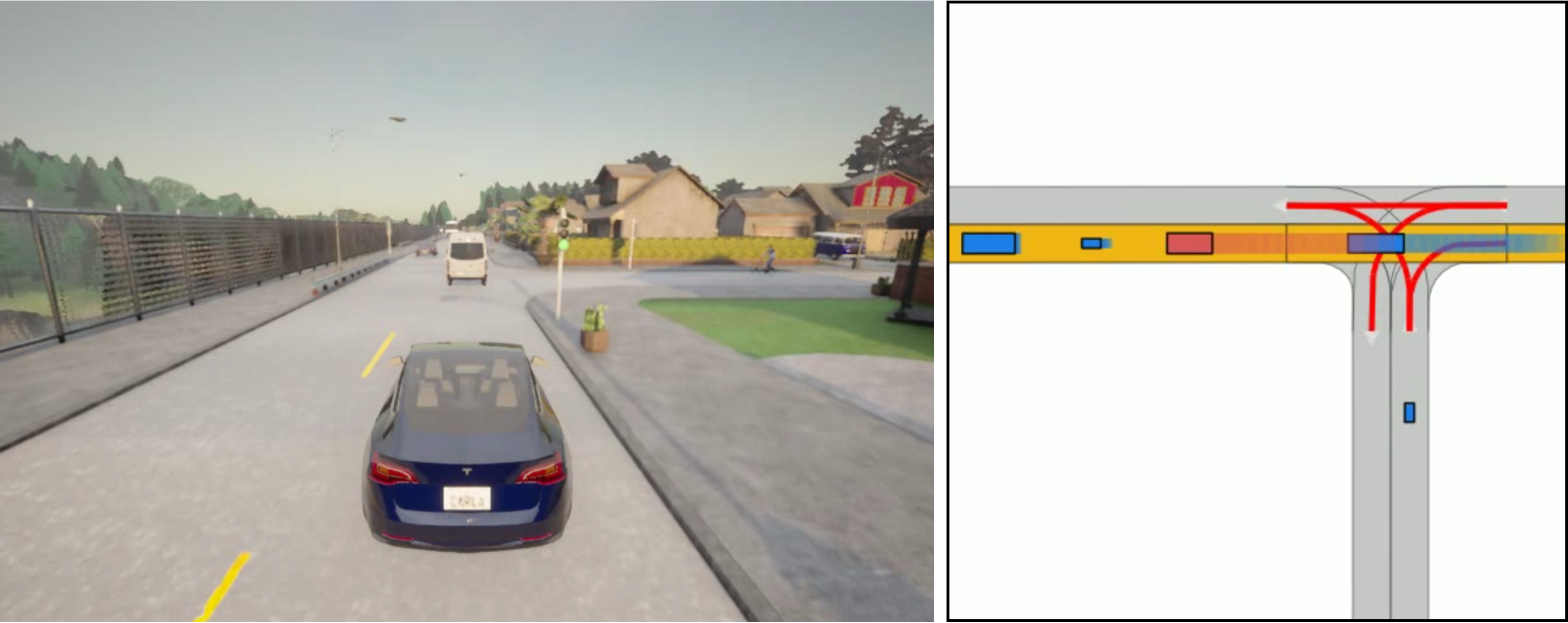}
	\caption{Snapshot from the CARLA simulator, where the corresponding motion planning problem is visualized on the right hand side.}
	\label{fig:CARLA}
\end{figure}

In contrast to CommonRoad scenarios which consist of a single planning problem, for the CARLA simulator \cite{Dosovitskiy2017} we use a navigation module to plan a route to a randomly chosen destination on the map. We then follow this route by replanning a trajectory with a duration of 3\si{\second} every 0.3\si{\second} until the vehicle reached the destination, where we combine our decision module with the optimization-based planner described in Sec.~\ref{subsec:commonRoad}. Moreover, since we use the high-fidelity vehicle model from CARLA for which the kinematic single track model in \eqref{eq:dynamics} is just an approximation, we additionally apply a feedback controller that counteracts model uncertainties and disturbances. 
For the experiments we consider the Town 1 map and use a constant velocity assumption to predict the future positions of the surrounding traffic participants. Tab.~\ref{tab:CARLA} displays the results for three different routes, and an exemplary snapshot from the CARLA simulator is shown Fig.~\ref{fig:CARLA}. The outcome demonstrates that our decision module performs very well as part of a full autonomous driving software stack consisting of navigation, prediction, decision making, motion planning, and control, and therefore enables robust motion planning in real-time.

\section{Conclusion}

We presented a novel approach for decision making in autonomous driving, which applies set-based reachability to identify driving corridors. As we demonstrated with an extensive numerical evaluation on 2000 CommonRoad traffic scenarios, our decision module runs in real-time, can be combined with multiple different motion planners, and leads to significant speed-ups compared to executing a motion planner standalone. Moreover, our experiments in the CARLA simulator, for which we integrated our decision module into a full autonomous driving software stack, showcase that our approach performs well for a lifelike setup close to reality. 

\bibliography{kochdumper,cpsGroup}
\bibliographystyle{IEEEtran}

\end{document}